\pdfoutput=1

\documentclass[11pt]{article}

\usepackage{acl}

\usepackage{times}
\usepackage{latexsym}
\usepackage{float}

\usepackage[T1]{fontenc}

\usepackage[utf8]{inputenc}

\usepackage{microtype}

\def\thick{\noalign{\hrule height 1pt}}

\usepackage{graphicx}
\usepackage{linguex}
\usepackage{soul}
\usepackage{xcolor}

\title{``No, they did not'': Dialogue response dynamics in pre-trained language models}

\author{Sanghee J. Kim$^1$, Lang Yu$^2$, Allyson Ettinger$^1$ \\
$^1$Department of Linguistics, University of Chicago \\
$^2$Meta \\
\texttt{\{sangheekim,aettinger\}@uchicago.edu}, \texttt{langyu@fb.com} \\ 
}

\begin{document}

\maketitle

\begin{abstract}
A critical component of competence in language is being able to identify relevant components of an utterance and reply appropriately. In this paper we examine the extent of such dialogue response sensitivity in \mbox{pre-trained} language models, conducting a series of experiments with a particular focus on sensitivity to dynamics involving phenomena of \mbox{at-issueness} and \mbox{ellipsis.} We find that models show clear sensitivity to a distinctive role of embedded clauses, and a general preference for responses that target main clause content of prior utterances. \mbox{However,} the results indicate mixed and generally weak trends with respect to capturing the full range of dynamics involved in targeting \mbox{at-issue} versus \mbox{not-at-issue} content. Additionally, models show fundamental limitations in grasp of the dynamics governing \mbox{ellipsis,} and response selections show clear interference from superficial factors that outweigh the influence of principled discourse constraints.
\end{abstract}

\section{Introduction}

Competence in language involves understanding complex principles governing relevance of previous content and dynamics of referring back to that content. Certain parts of an utterance are more central and more likely to receive a response than others, and the pragmatic and grammatical rules governing responses in dialogue interact with the nature of the content being responded to. 
Humans are highly sensitive to these distinctions, and we can expect these sensitivities to be critical for robust models in NLP, and especially for dialogue. 
Here we examine sensitivity to these dialogue response dynamics in pre-trained language models (PLMs). PLMs are now used as foundation for nearly every downstream NLP task, including dialogue applications~\citep[e.g.,][]{upadhye2020, koto2021discourse}. The impressive downstream performance enabled by these models has raised important questions about what types of linguistic competence are being learned during pre-training---and though there is a growing body of work answering aspects of this question, topics of pragmatic and dialogue competence have been relatively understudied. In this paper we focus on addressing this gap, and in particular on understanding the extent to which PLMs develop sensitivity to dynamics governing responses in dialogue. Though these PLMs are not trained to engage in dialogue per se, they can be expected to encounter dialogue during training (in, for instance, novels), so it is not unreasonable to expect that they may learn about such dialogue dynamics along with other linguistic competences. The strength of these models' sensitivity to such dynamics has important implications for robustness in dialogue applications, since a strong grasp of dialogue dynamics in standard PLMs stands to reduce fine-tuning needs and enable more robust downstream behaviors. 

We begin with the notion of \emph{at-issueness}. A component of an utterance is considered \emph{at-issue} if it is part of the ``main point'' of the utterance---this is to be contrasted with side comments or mentions of background knowledge, which are not the main focus of the sentence. As we lay out in Section~\ref{sec:atiss-background}, the distinction between at-issue and not-at-issue content of an utterance is reflected directly in the nature of responses to that utterance. We thus examine models' preferences for different responses, to assess whether the preferences reflect understanding of at-issueness and how to respond to it. We find that models show consistent preference to target at-issue (main clause) content, but mixed and overall fairly weak sensitivity when it comes to the full range of dynamics involved with at-issueness.

These assessments of at-issueness sensitivity are also critically reliant on another aspect of dialogue response dynamics: ellipsis. We thus additionally make a closer examination of the extent to which constraints from context dictate models' selection of auxiliary verbs (such as \emph{did, does, would}) in ellipsis constructions. We find that although models often favor an auxiliary verb that targets the main clause, they also make frequent errors, and they very rarely favor both of the auxiliary forms that align with the prior context. These results furthermore raise the important possibility that models are highly sensitive to preferences for particular auxiliary verb types, and that this could drive the at-issueness results as well. With this in mind we revisit the at-issueness experiments, and find that, indeed, there are substantial differences in models' preferences depending on the identity of the particular verb that targets the relevant content.

Overall, our results suggest that PLMs have non-trivial gaps in their understanding of response dynamics in dialogue. Our results also indicate certain differences between models: BERT and RoBERTa show strong bias toward selecting responses that target the most recent and/or main clause content, while other models show more reliance on individual auxiliary verb properties. In all cases the results indicate that these PLMs have not yet achieved ideal sensitivity to response dynamics involving at-issueness and ellipsis, and that effectiveness in dialogue will benefit from additional training approaches. We make all datasets and code available for further testing.\footnote{\url{ https://github.com/sangheek16/dialogue-response-dynamics}}

\section{Related work}

Recent years have seen extensive work on analysis of PLMs. Methodologically, some of the most popular analysis paradigms targeting model embeddings have included classification-based probing~\cite[e.g.,][]{kim2019probing, paws2019naacl} and correlation with similarity judgments \cite{finkelstein2001placing, gerz2016simverb, conneau2018senteval}. Other work has analyzed PLMs by eliciting and analyzing output predictions~\cite{linzen2016assessing,goldberg2019assessing}. Our work here focuses primarily on the latter methodology, examining and comparing model output probabilities---however, our analysis in Section~\ref{sec:probing} uses classification-based probing. Our work also builds on approaches implementing specialized sentence generation systems that produce large annotated datasets~\cite{ettinger2018assessing,mccoy2019right}. 

Analyses of PLMs have targeted a variety of types of linguistic competence. In particular, a large body of work has studied the extent to which PLMs capture syntactic and semantic information~\cite{linzen2016assessing,peters2018dissecting,bacon2019does,hewitt2019structural, tenney2019you}. Less work has addressed the extent to which PLMs show sensitivity to pragmatic and discourse information, as we focus on in this paper. \citet{kurfali2021probing} study multilingual models in various discourse tasks via zero-shot learning. \citet{pandia2021pragmatic} investigate LMs' pragmatic competence to predict discourse connectives. \citet{pitler2009using} report that a supervised classifier is able to identify discourse relations given syntactic features along with connectives. \citet{patterson2013predicting} implement a similar idea and show that classifiers are able to predict the presence of a connective based on shallow linguistic cues. \citet{koto2021discourse} explore pre-trained language models' capability in capturing discourse level relations. We complement this existing work by branching into new areas of pragmatic and discourse knowledge, examining models' sensitivity to dialogue response dynamics.

Another closely related literature is that in which PLMs, especially transformer LMs, are used for building dialogue systems directly. \citet{le2019multimodal} propose Multimodal Transformer Networks (MTN) for visual-grounded dialogue tasks. Other work investigates topic-driven language models for emotion detection in dialogues~\cite{zhu2021topic}. \citet{oluwatobi2020dlgnet} report state-of-the-art performance on dialogue generation using transformer-based models. 
There are also language models designed for and trained on dialogue or conversation, 
such as 
TransferTransfo \citep{wolf2019transfertransfo}, 
PLATO \citep{bao2020plato},
ConveRT \citep{henderson2020convert}, TOD-BERT \citep{wu2020tod}, DialoGPT \citep{zhang2020dialogpt}, DialogBERT \citep{gu2021dialogbert}, and LaMDA \citep{thoppilan2022lamda}.

Here we focus on clarifying the extent to which PLMs pre-trained in the standard paradigm can develop knowledge of dialogue dynamics prior to any specialized dialogue training. This line of inquiry serves to broaden our general understanding of linguistic competence of standard PLMs, and also has implications for use of these standard PLMs as foundation for further dialogue-specific training.

\vspace{-0.8em} 
\nopagebreak 

\section{Background}

\subsection{At-issueness}\label{sec:atiss-background}

Our analyses focus on the dynamics that govern responses in dialogue, and aspects of prior utterances that they target. The first notion that we test for in PLMs is sensitivity to ``at-issueness.'' At-issueness refers to content's status as the main point of the utterance---to be contrasted with not-at-issue content, such as side comments and assumed knowledge (see \citet{potts2005} for a comprehensive overview). 
Humans are sensitive to which content in an utterance is ``at-issue'' and which content is not---and this sensitivity is reflected in dialogue response dynamics. Consider the utterance in \ref{ex:arc}.

\ex.\label{ex:arc} The nurse, who has interest in French cuisine, adopted a rescue dog.

If a listener responds to \ref{ex:arc} with ``No'' or ``That's not true,'' they would most likely be objecting to the claim that \emph{the nurse adopted a rescue dog}, since this is the main point (at-issue content) of \ref{ex:arc}. It is less likely that they would be objecting to the side comment about French cuisine. As a result, a response of ``No, he didn't (adopt a rescue dog),'' would be natural, while ``No, he doesn't (have interest in French cuisine)'' would be less so. 

This intuition drives a key diagnostic used to distinguish at-issue and not-at-issue content, known as the \textbf{Rejection \& Peripherality Test} (or the Assent/Dissent Test) \citep{amaral2007,koev2013,syrett2015}. The ``rejection'' component of this test is illustrated in \ref{ex:rejection}. Speaker B$_1$ replies to Speaker A's utterance with a \textbf{rejection} (``No''), and uses the elliptical verb phrase (``did not'') that targets the (at-issue) content of the main clause (``The nurse adopted a rescue dog.''), for a natural and appropriate response. In contrast, Speaker B$_2$ rejects the (not-at-issue) content inside the appositive relative clause (ARC), which is less natural (indicated with `\#').

\ex. \label{ex:rejection}
\a. Speaker A: ``The nurse, who has interest in French cuisine, adopted a rescue dog.''
\b. \label{ex:rejection-a}Speaker B$_1$: ``No, he did not.'' [Targeting at-issue content]
\c. \label{ex:rejection-n}Speaker B$_2$: {\#}``No, he does not.'' [Targeting not-at-issue content]

There is, however, a more natural way to object to not-at-issue content: pausing the dialogue to question a side comment or assumption. This is highlighted in the \textbf{peripherality test}, which uses phrases like, ``Hey, wait a minute'' \citep{von2004,amaral2007}, or ``Wait, this is peripheral to your point but...'' \citep{koev2018} in order to make targeting not-at-issue content more acceptable. We show an example in \ref{ex:peripherality}.

\ex. \label{ex:peripherality}
\a. Speaker A: ``The nurse, who has interest in French cuisine, adopted a rescue dog.''
\b. \label{ex:peripherality-n}Speaker B: ``Wait no, he does not (have interest in French cuisine).'' [Targeting not-at-issue content]

Human sensitivity to this pattern of relationship between at-issueness and ``No'' versus ``Wait no'' response types has been well attested in psycholinguistic experiments. \citet{syrett2015} in their Experiment 1 find that when selecting between responses that target not-at-issue content in an embedded clause of a prior utterance, humans are much more likely to choose a response of type ``Wait no'' (77\%) than of type ``No'' (23\%).\footnote{The specific wordings in this experiment were ``Hey, wait a minute,'' and ``That's not true.''} By contrast, when selecting between responses that target at-issue content in a main clause of a prior utterance, humans' rate of selection of these two response types is roughly even. In their Experiment 2, \citet{syrett2015} furthermore show that when selecting among ``No'' type responses, humans have a strong preference for choosing those that target at-issue content of prior utterances (73.9\%) compared to not-at-issue content (26.1\%).

Leveraging this knowledge of human sensitivities, we make use of diagnostics modeled after the Rejection \& Peripherality Test to examine whether PLMs are also sensitive to these discourse dynamics involving at-issueness and response type. For structuring not-at-issue content, we focus on ARCs as used in the examples above.

\subsection{Ellipsis}

The examples above make critical use of the grammatical phenomenon of ellipsis: use of abbreviated verb phrases that refer back to previous verb phrases. In ellipsis, typically an auxiliary verb (like \emph{did, does, would}) remains as the verb in the elided verb phrase---for instance: ``No, he didn't'' is an elided form that could refer back to ``The nurse adopted a rescue dog,'' standing in for the longer phrase ``No, he didn't adopt a rescue dog.'' Ellipsis is another critical component of forming responses in dialogue, and it plays an important prerequisite role in assessing at-issueness. For these reasons, we also test models' grasp of ellipsis in dialogue.

\section{Experiments} 

\subsection{Construction of test items}

To enable controlled tests inspired by the structure of the Rejection/Peripherality tests, we generate items using templates. Each input item consists of a sequence of two sentences: (a) a \textbf{context sentence}, and (b) a \textbf{response sentence}. 

We generate the \textbf{context} sentences based on a core template of ``\textsc{NounPhrase}, who \textsc{VerbPhrase1}, \textsc{VerbPhrase}2.'' This structure includes an embedded ARC (not-at-issue content) and an embedding main clause (at-issue content), as in our example \ref{ex:arc} above: \emph{The nurse, who has interest in French cuisine, adopted a rescue dog.}
For the noun phrases, we sample from a list of nouns referring to names of occupations (e.g., \emph{nurse, reporter, violinist}). As for verb phrases, to ensure that it would always be unambiguous whether a rejection is targeting the main or the embedded clause, for each item we control the verb phrases of the two clauses such that they will always be targeted by different elided verbs in the response sentence.
To do this we create ordered verb pairs from six unique auxiliary verbs \emph{is, was, does, did, has, could}, with the first verb assigned to the embedded clause, and the second assigned to the main clause. This resulted in 30 (=$_{6}P_{2}$) unique verb type pairings. We then draw from a list of verb phrases associated with each auxiliary verb: for instance, the verb phrases for the auxiliary verb \emph{does} contain examples such as \emph{has interest in French cuisine}, and \emph{enjoys hiking}; for the verb \emph{did}, the verb phrases include \emph{adopted a rescue dog}, and \emph{met the Illinois governor at a Greek restaurant.} We randomly sample from these verb phrases for each verb pair, with the phrase for the first verb assigned to the \textsc{VerbPhrase1} position in the template, within the ARC, and the second verb phrase to the \textsc{VerbPhrase2} position, in the main clause. 
Ten unique sentences were generated for each verb pair, resulting in 300 context sentences (= 30 verb pairs {*} 10 sentences). Because of our use of the ordered pairs, every auxiliary verb is equally likely to be the correct form for targeting either at-issue content or not-at-issue content of a context sentence.

The \textbf{response} sentences then include a ``header'' consisting of either \emph{No} or \emph{Wait no}, a subject pronoun (sampled randomly to avoid pronoun gender biases), an auxiliary verb targeting either the main clause or embedded clause, and \emph{not}. For example, response sentences might consist of ``No, she does not,'' or ``Wait no, he has not.'' The differences in these headers are the critical factor that impacts whether a response sentence can reasonably target (not-)at-issue content in the context sentence---and the auxiliary verb indicates which verb phrase in the context sentence is being targeted. 

In constructing these items, an additional consideration is how to create a setting in which the PLMs may naturally recognize the input as describing a dialogue. We choose to present the items in a format of dialogue resembling that in novels, where entities are described explicitly as uttering the relevant statements. Our final templates thus take a form as shown in example \ref{ex:item} below.\footnote{We also tested with a simpler dialogue style: \emph{A: ``The nurse, who has interest in French cuisine, adopted a rescue dog.'' B: ``\{No / Wait no\}, he \{did / does\} not.}'' This formatting difference did not significantly change the results in the experiments for which we made this comparison.} We randomly sample the speaker names (e.g., Marco, Ellie) from a list of 400 names, ensuring that no two names repeat in a given item.

\ex. \label{ex:item} Marco said, ``The nurse, who has interest in French cuisine, adopted a rescue dog,'' and Ellie replied, ``\{No / Wait no\}, he \{did / does\} not.''

\subsection{Models tested} 
In all of our experiments below, we test six PLMs. Of these models, five are masked language models (MLMs): BERT \citep{devlin2018bert}, RoBERTa \citep{liu2019roberta}, XLM-RoBERTa \citep{conneau2020unsupervised}, DistilBERT and DistilRoBERTa~\cite{sanh2019distilbert}.
The final model is a causal (unidirectional) language model (CLM): DistilGPT2 \cite{distilgpt2}. 
We used the implementations of these models made available through the HuggingFace Transformers library \citep{wolf2020transformers}.

\section{At-issueness tests}\label{sec:atissue}

\subsection{Header preferences}\label{sec:header}

We begin by asking whether models, like humans, are sensitive to the role of the response ``header'' (``No'' vs. ``Wait no'') in whether a rejection can naturally target (not-)at-issue content. 
In line with Experiment 1 in \citet{syrett2015}, we begin by testing whether models recognize that ``No'' is an appropriate header when the response auxiliary targets the main clause, but ``Wait no'' is critically more appropriate when the response auxiliary targets the embedded clause. To do this, for a given item we hold constant the auxiliary verb in the response sentence (e.g., \emph{did/does/has}), and we compare the model probabilities for headers of ``No'' vs ``Wait no.'' The auxiliary verb for a given item either targets the main clause (at-issue) content of the context sentence, or targets the embedded clause (not-at-issue) content. Since these items are different lengths depending on the choice of header, we compare the conditional log probability of the full sequence, normalized by length, for both MLM and CLM models. For MLMs, we compute pseudo-log-likelihoods, which are obtained by summing the conditional log probabilities of each sentence token (as in \citet{salazar2020}), and normalizing by number of input tokens.\footnote{We use the minicons library \citep{misra2022minicons} for conditional sequence probabilities for all models.}

\begin{figure}[t]
\center
\includegraphics[width=0.48\textwidth, height=4.3cm]{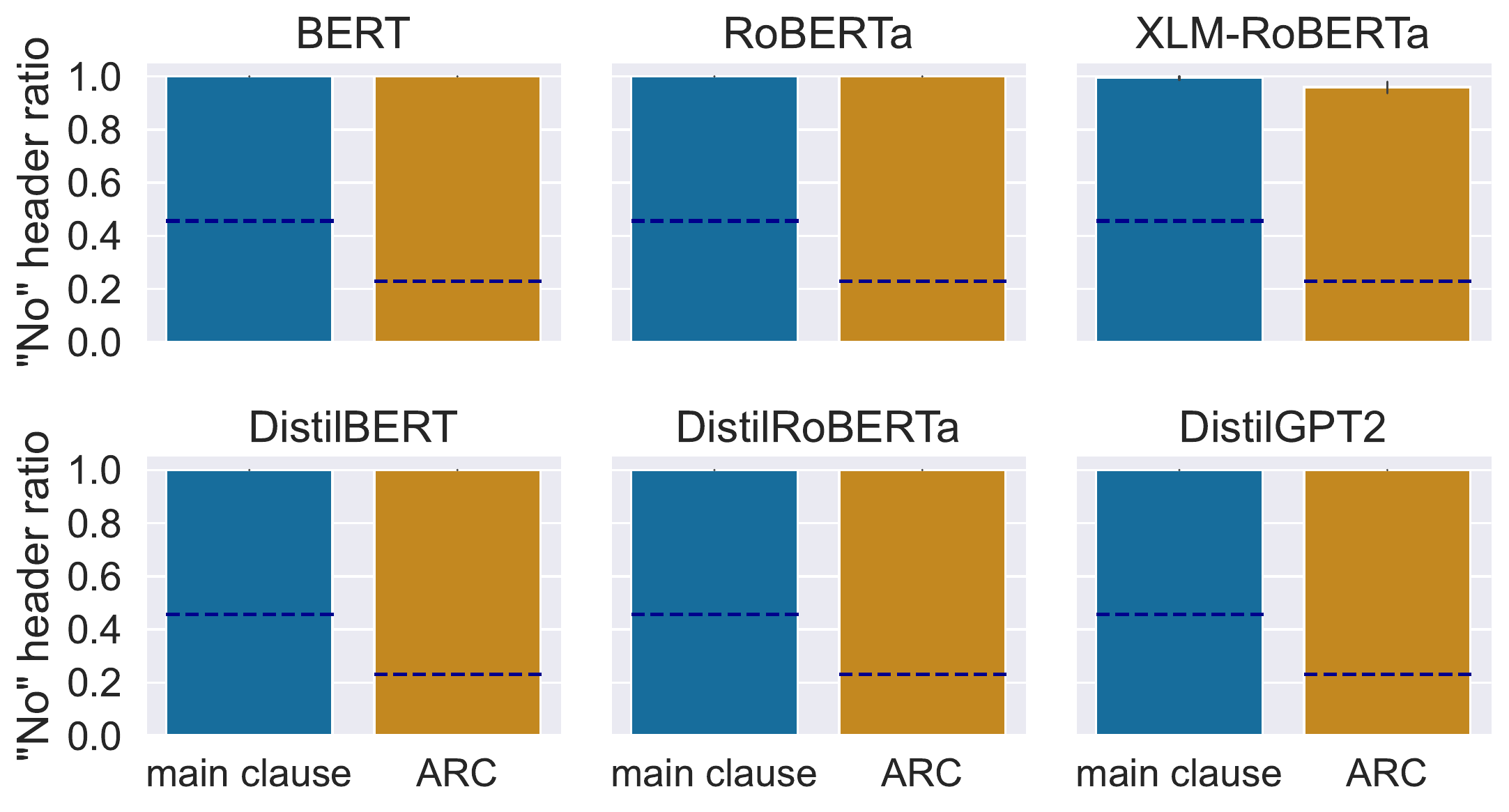}
\caption{\label{atissue-header} Header selection. The Y-axis indicates the ratio of the ``No'' header having higher probability than the ``Wait no'' header given the same verb phrase in the target sentence. The X-axis indicates the target content: `main clause' condition is when the main clause is targeted; `ARC' condition is when the embedded clause is targeted. Dashed lines are human performance baseline reported in \citet{syrett2015} (Experiment 1). Error bars = 95\% Confidence Interval.}
\end{figure}

Figure \ref{atissue-header} shows the percentage of items for which the model assigns a higher probability to the sequence with ``No'' than with ``Wait no,'' separated based on whether the response auxiliary verb targets the main clause (at-issue content) or embedded clause (not-at-issue content). We see that regardless of which clause the response auxiliary targets, models always prefer the ``No'' header to the ``Wait no'' header, in a strong contrast with humans' intuition that ``Wait no'' is much better for targeting not-at-issue content.

\subsection{Comparing auxiliary preferences}\label{sec:rejection}

While the above result casts doubt on models' sensitivity to the relationship between at-issueness and response headers, we might wonder whether ``No'' is simply too strong or too frequent a response. Additionally, the fact that ``No'' and ``Wait no'' are different in lengths raises questions about whether the probability comparisons are fair in the MLMs (in which use of full sentence probabilities is also a bit atypical). 

To explore a different angle on this question, we therefore shift to a direct examination of models' preferences for response auxiliaries that target the at-issue content, versus those that target not-at-issue content---and how this is affected by the nature of the ``No'' versus ``Wait no'' header. This allows us to examine the MLMs in a more natural setting (assessing probabilities on a single masked position), and also allows us to examine how headers impact models' choices for what contextual content should be targeted.

In these experiments, for MLMs we simply place a [MASK] token at the response auxiliary position and compare auxiliary probabilities at that position:

\ex. \label{ex:item-mlm} Marco said, ``The nurse, who has interest in French cuisine, adopted a rescue dog,'' and Ellie replied, ``\{No / Wait no\}, he [MASK] not.''

For the CLM, we compare probabilities for the full sequence, with one of two candidate auxiliaries in the target position (as in the previous experiment). The two candidate auxiliaries that we insert are simply the two most relevant: the auxiliary that targets the main clause (at-issue content), and the one that targets the embedded clause (not-at-issue content).

\begin{figure}[t]
\center
\includegraphics[width=0.47\textwidth]{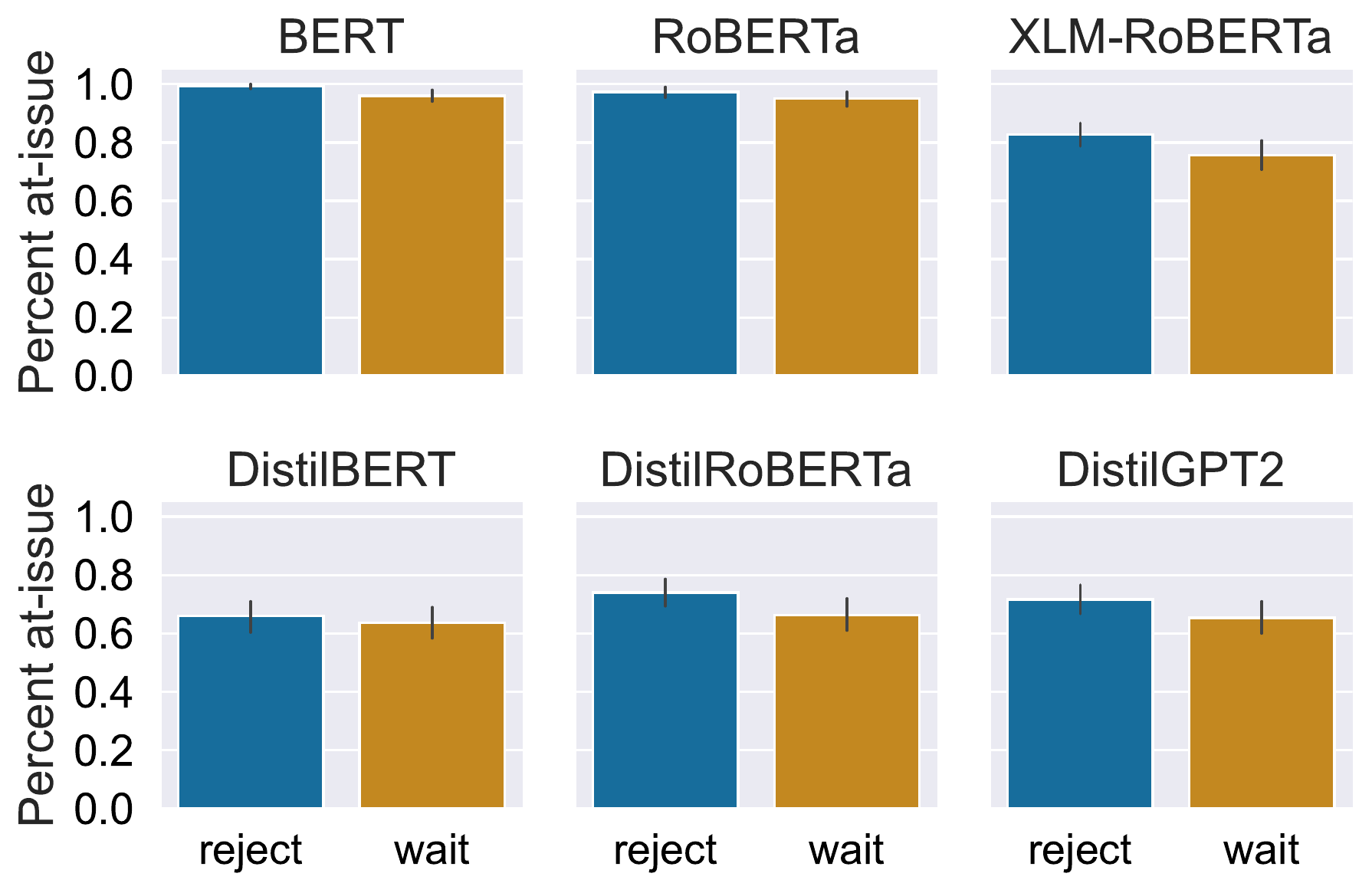}
\caption{\label{atissue}Rejection test. The Y-axis indicates the percentage of instances in which models show preference for responses targeting at-issue content over not-at-issue content. The X-axis indicates type of header in the input sequence: `Reject' = ``No'' header; `Wait' = ``Wait no'' header. Error bars = 95\% Confidence Interval.}
\end{figure}

Figure \ref{atissue} shows the percentage of the time that each model assigns higher probability to the auxiliary targeting the at-issue content, over the auxiliary targeting the not-at-issue content.\footnote{Based on the human performance in \citet{syrett2015} (Experiment 2), we could expect a humanlike ratio of selecting at-issue content with the ``No'' header to be 0.789. This version of the human experiment did not obtain a ratio for the ``Wait no'' header.} We see that all models prefer the at-issue-targeting auxiliary at a rate greater than chance, with some models (BERT, RoBERTa) showing preference for targeting the at-issue content almost 100\% of the time. The question, then, is whether the use of ``Wait no'' reduces the rate of targeting the main clause---given that this header allows for targeting of not-at-issue content. A one-sided $t$-test shows that the selection ratio of the at-issue content is indeed larger with the ``reject'' header compared to the ``wait'' header in most of the tested models, at reasonable levels of statistical significance (BERT: $t$ = 2.716, $p$=0.003; RoBERTa: $t$ = 1.489, $p$ = 0.069; XLM-RoBERTa: $t$ = 2.115, $p$ = 0.017; DistilBERT: $t$ = 0.597, $p$ = 0.275; DistilRoBERTa: $t$ = 2.056, $p$ = 0.02; DistilGPT2: $t$ = 1.671, $p$ = 0.048). This suggests that at least some of the models may have picked up on some relationship between these headers and targeting of at-issue versus not-at-issue content, though it is also clear that these trends are relatively weak.

\subsection{Conjunction}\label{sec:conjunction}
The above results show that PLMs exhibit a strong preference for auxiliaries that target the main clause of the context sentence. How should we interpret this preference for targeting the main clause? An immediate question that arises is whether this preference could be due to recency alone: in our items, the verb phrase in the main clause of the context is also always the more recent verb phrase before the response sentence. To investigate this possibility, we modify our items to involve two verb phrases in the context sentence, but with the phrases joined by conjunction \ref{ex:item-conj}. This means that both verb phrases are now at-issue, and any preference for one over the other can be attributed to recency.

\ex. \label{ex:item-conj} Marco said, ``The nurse has interest in French cuisine and adopted a rescue dog,'' and Ellie replied, ``\{No / Wait no\}, he [MASK] not.''

\begin{figure}[t]
\center
\includegraphics[width=0.47\textwidth, height=4.9cm]{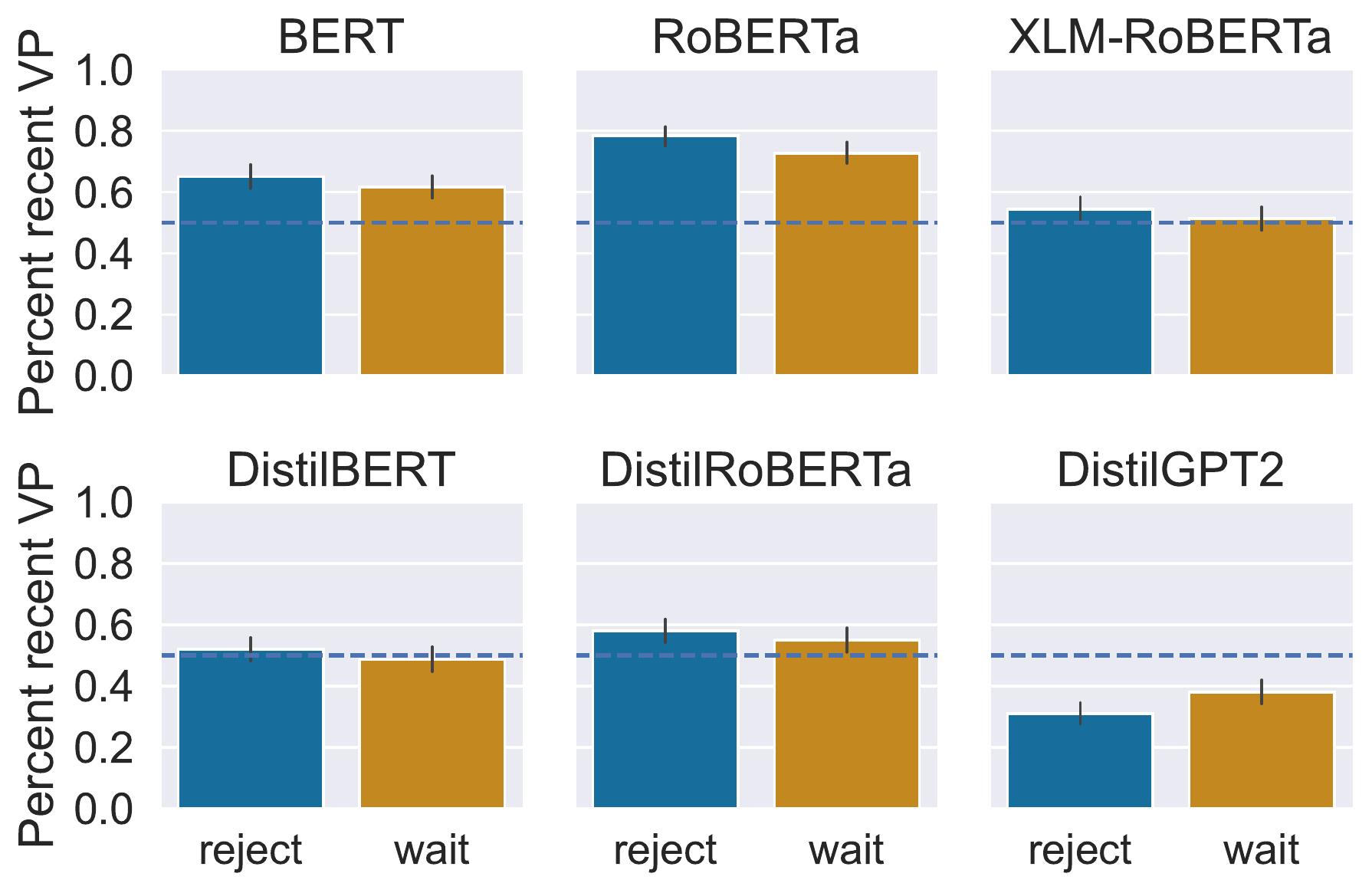}
\caption{\label{conjunction}Conjunction test. 
The Y-axis indicates percentage of instances in which models show preference for responses targeting more recent verb phrases over more distant verb phrases. The X-axis indicates type of header in the input sequence: `Reject' = ``No'' header; `Wait' = ``Wait no'' header. Dashed lines mark chance level (50\%). Error bars = 95\% Confidence Interval.}

\end{figure}

We again compare model probabilities with each of the two valid candidate auxiliaries. Figure~\ref{conjunction} shows the percentage of items for which the models prefer the auxiliary that targets the more recent verb phrase in the context sentence. It is clear from these results that the trend toward targeting main clause content in Figure~\ref{atissue} cannot be attributed to recency alone: a majority of models are now hovering around 50\% in targeting the most recent phrase, with DistilGPT2 in fact showing a preference to target the more \emph{distant} phrase rather than the more recent one. BERT and RoBERTa, by contrast, do both show some bias to target the more recent verb phrase---however, this trend is substantially weaker than the trend in Figure~\ref{atissue}, indicating that although these models do prefer to target more recent content, they also show a preference for targeting main clause content over and above this recency bias.

\subsection{Probing} \label{sec:probing}

The results above suggest that at very least, models are sensitive to the fact that embedded clauses (in this case ARCs) have special status in affecting response dynamics---such that models prefer response auxiliaries that target the main clause in the previous context, over and above effects of recency. In this section we briefly confirm that models are sensitive to the differing status of embedded clause content, through a probing experiment testing whether model representations distinguish embedded clause (not-at-issue content) from main clause (at-issue content). To do this, we extract token embeddings from the models and train a classifier to predict whether these tokens are part of a not-at-issue content or an at-issue content. The task is formulated as 3-class classification: contextualized token embeddings from the last hidden layer are used as input for the classifier, and labels are generated based on whether the token is 1) part of at-issue content, 2) part of not-at-issue content or 3) neither.\footnote{Tokens counted as ``neither'' are those like ``Marco said'' that are used to introduce the dialogue content.} 
Train/test dataset are randomly split for each model, while keeping tokens from the same input sequence together, yielding on average 8,500 training and 4,000 test samples.\footnote{To mitigate impacts of random variation in train/test split across models, we trained the probe for each model three times and averaged the results.} For this experiment we use a multi-layer perceptron classifier with a single hidden layer of size 50 with ReLU activation, and a softmax layer to generate ternary labels. We use a relatively simple classifier following the reasoning of~\citet{adi2017fine}, that this allows examination of how easily extractable information is in these representations.

\begin{table}[t!]
\centering
\resizebox{0.31\textwidth}{!}{
\begin{tabular}{c|c}
\thick
\textbf{Model}      & Accuracy (\%) \\ \thick
BERT       &    99.9      \\ \hline
RoBERTa           &   100      \\ \hline
XLM-RoBERTa         &  99.2     \\ \hline
DistilBERT &   99.4       \\ \hline
DistilGPT2 &   99.5             \\ \hline
DistilRoBERTa & 100         \\ \thick
\end{tabular}
}
\caption{Probing performance using token embeddings from last hidden layers.}
\label{tab:probing}
\end{table}

As shown in Table~\ref{tab:probing}, all models achieve near perfect classification accuracy. The result further supports the conclusion that these models do encode distinctions between content in the main clauses of these sentences and content in embedded clauses---such that the trends in favor of targeting main clause content may be considered to reflect some real sensitivity to contributions of these structural properties to dialogue dynamics.

\section{Ellipsis}
The response tests above rely on a critical prerequisite: that models understand how to use ellipsis structures like ``he didn't'' and ``she doesn't.'' In the case of our items, it is specifically the case that there are only two auxiliary verbs that could possibly be appropriate in a given response sentence, because there are only two verb phrases in the context sentence that could be rejected. In this section we take a closer look at whether models' preferences for response auxiliaries reflect these broader discourse constraints on ellipsis. 

\subsection{Ellipsis top one accuracy}\label{sec:ellipsis-top1}

We begin by examining the auxiliaries that receive top probability from the models, among the six tested auxiliary verb candidates (i.e., \textit{did}, \textit{does}, \textit{has}, \textit{is}, \textit{was}, and \textit{would}). Specifically, we ask whether the highest-probability response auxiliary selected by the model for a given context is appropriate given the context sentence and header. This test differs from our comparisons above because the previous tests simply compared the two relevant auxiliaries (main clause and embedded clause), without testing whether either of these auxiliaries was assigned the highest probability among all possible auxiliary verbs. Here we count the model as correct if in the case of the ``No'' header it assigns the highest probability to the auxiliary that targets the main clause, or if in the case of the ``Wait no'' header it assigns the highest probability to either the main-clause-targeting or embedded-clause-targeting auxiliary (because ``Wait no'' could also reasonably target the at-issue content).

\begin{figure}[t]
\center
\includegraphics[width=0.48\textwidth]{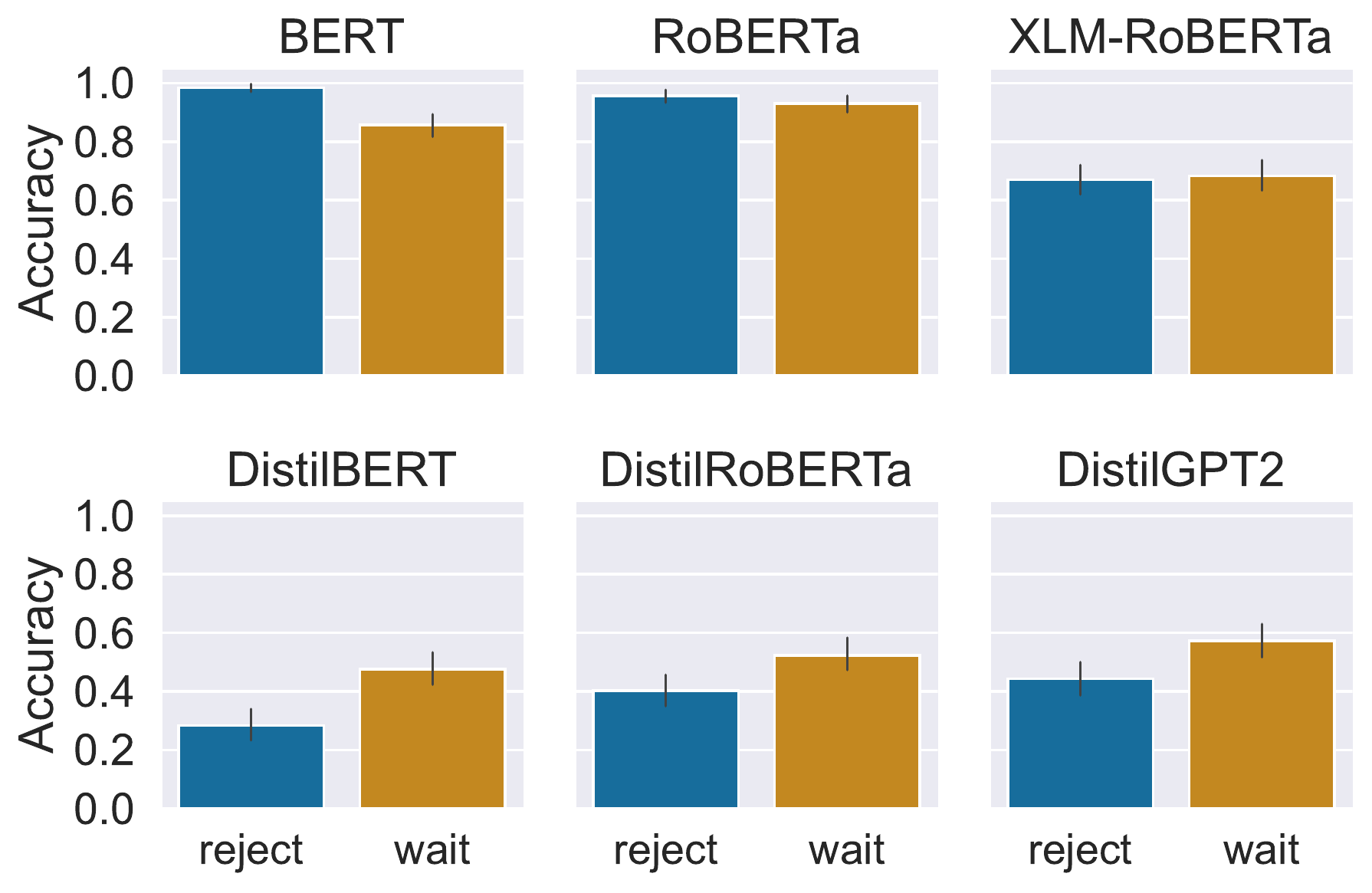}
\caption{\label{ellipsis-accuracy}Ellipsis test (top-1 accuracy). The Y-axis indicates the percentage of instances in which models prefer responses targeting the main clause (in the case of the ``No'' header), or targeting either the main clause or embedded clause (in the case of the ``Wait no'' header). The X-axis indicates type of header in the input sequence: `Reject' = ``No'' header; `Wait' = ``Wait no'' header. Error bars = 95\% Confidence Interval.}
\end{figure}

Figure \ref{ellipsis-accuracy} shows the percentage of the time that the top-ranked auxiliary is among those counted as correct based on the header. We see wide variation in the models' performance on this assessment, with BERT and RoBERTa preferring the correct auxiliary nearly 100\% of the time, but distilled models rarely selecting the correct auxiliary as top choice. This suggests at first glance that BERT and RoBERTa have gained a stronger grasp on the relationship of elided auxiliary forms to the previous context---however, it must also be noted that our definition of ``correct'' favors BERT and RoBERTa because preference for targeting the main clause (which these two models have exhibited) can always be counted as correct. We thus implement a more difficult ellipsis test in the next section.

\subsection{Ellipsis top two accuracy}\label{sec:ellipsis-top2}

As we describe above, because there are only two verb phrases in each of our context sentences, it is clear that there are only two acceptable auxiliary verb forms that can occur in a given response sentence. To test whether models have a grasp of this constraint, in this section we examine the top two highest-probability auxiliaries, and assess the percentage of the time that these auxiliaries are exactly the two that target the main clause and embedded clause of the context sentence, respectively.

\begin{figure}[t]
\vspace{1em}
\center
\includegraphics[width=0.48\textwidth]{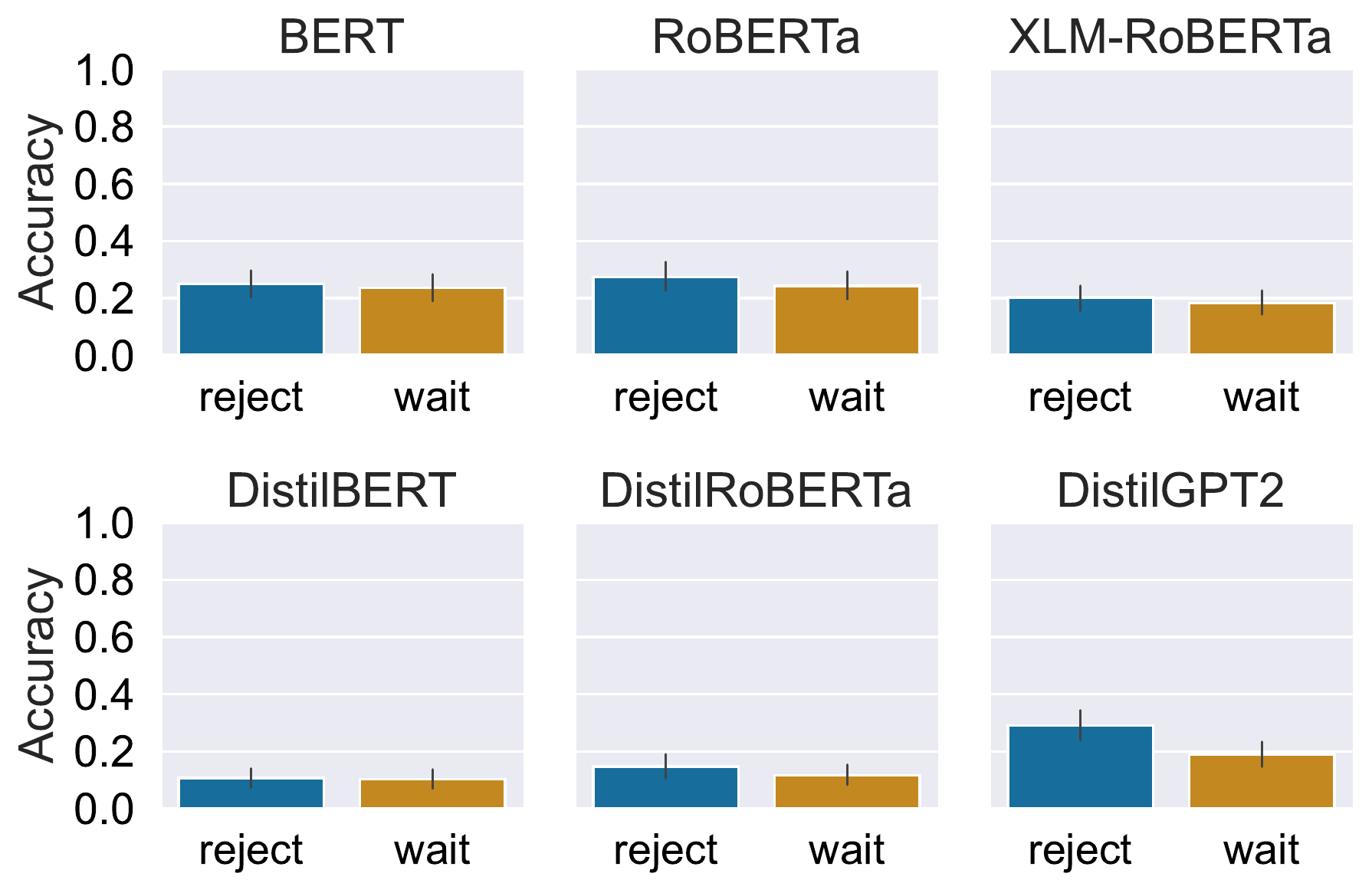}
\caption{\label{ellipsis-top2} Ellipsis test (top-2 accuracy). The Y-axis indicates the percentage of instances in which models' top two highest-probability auxiliary verbs are the two acceptable auxiliary verbs given context. The X-axis indicates type of header in the input sequence: `Reject' = ``No'' header; `Wait' = ``Wait no'' header. Error bars = 95\% Confidence Interval.}
\end{figure}

Figure \ref{ellipsis-top2} shows the percentage of the time that the two acceptable auxiliaries are the top two highest-probability auxiliaries for the models. It is clear that the accuracies here are very low---even the most accurate models meet the criterion only 20-30\% of the time, suggesting that this category of grammatical/discourse sensitivity is still largely missing from these models.

\subsection{Error analysis}

To get a better sense of where the models are going wrong in these tests, we perform error analyses for both of the two ellipsis tests. For the top-1 ellipsis test we examine cases where the top auxiliary is not ``correct,'' and for the top-2 test we examine cases where at least one inappropriate auxiliary ``intrudes'' in the model's top two. In Figures \ref{ellipsis-errorProportion-top1} and \ref{ellipsis-errorProportion-top2} in the Appendix,
we show the distribution of auxiliary verbs that the models prefer among these erroneous cases.
We see in these figures particularly substantial interference from more frequent auxiliaries like \textit{did}, \textit{does}, and \emph{is}, suggesting that rather than guiding auxiliary choice based primarily on discourse constraints, the model probability distributions are non-trivially influenced by general frequency of the individual auxiliary verbs in ellipsis.\footnote{Among the possible verb phrase ellipsis triggers (e.g., \emph{be}, \emph{has}, \emph{do}, etc.), the auxiliary verb \emph{do} has been reported to be the most frequent (44\%), with the auxiliary verb \emph{be} following the next (22\%) \citep{bos2011annotated}.}

\begin{figure*}[h!]
\center
\includegraphics[width=1.01\textwidth, height=5.9cm]{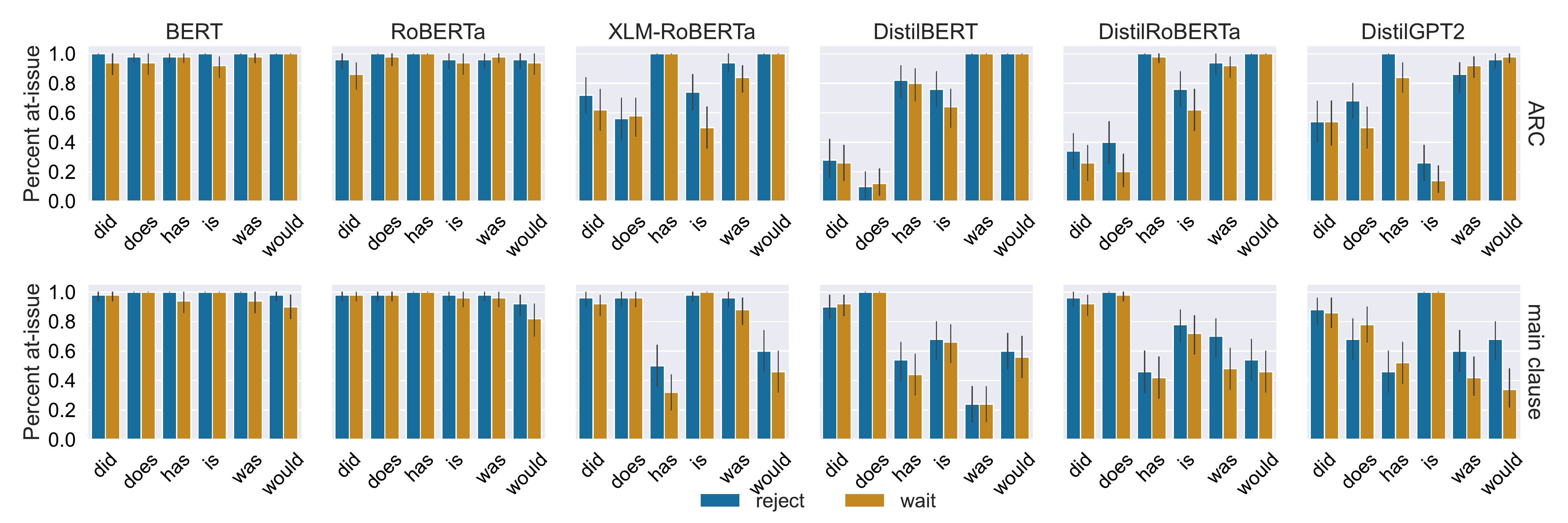}
\caption{\label{atissue-verb}Verb analysis on rejection test. The Y-axis indicates the percentage of instances in which models show preference for responses targeting at-issue content over not-at-issue content. The X-axis indicates the identity of the response auxiliary that can be used to target the selected clause of the context sentence: in the top row the X-axis indicates the auxiliary that targets content inside the ARC (not-at-issue content) of the context sentence, and in the bottom row the X-axis indicates the auxiliary that targets content inside the main clause (at-issue content) of the context sentence. `Reject' = ``No'' header; `Wait' = ``Wait no'' header. Error bars = 95\% Confidence Interval.}
\end{figure*}

\section{Verb analysis in rejection test}

The previous section raises questions about the extent to which these PLMs have a grasp on the basic discourse constraints that govern ellipsis in response utterances---and this ellipsis serves as critical foundation for our at-issueness response tests in Section \ref{sec:rejection}. 
In particular, the error analysis above indicates that model probabilities are influenced in large part by biases in favor of particular frequent auxiliary verbs. In this section we thus return to our at-issueness test to examine behaviors of individual auxiliary verbs separately.

Figure \ref{atissue-verb} shows the percentage of the time that the models prefer targeting at-issue content, broken down by which auxiliary verb targets the ARC (top row) or the main clause (bottom row). We see that for four of the models, the identity of the auxiliary verb makes a substantial difference: for instance, distilled models strongly prefer to target the main clause if ``did'' or ``does'' is the auxiliary that targets the main clause---but if ``did'' or ``does'' targets the embedded clause, these models strongly prefer targeting the embedded clause. In other words, the distilled models appear in large part simply to be biased toward preferring ``did'' or ``does'' in the response sentence. BERT and RoBERTa do not show as much verb-specific fluctuation, instead preferring to target the main clause content regardless of which auxiliary verb does so. This suggests that these two models have a relatively more robust grasp on use of ellipsis to target particular clauses in previous context. As for sensitivity to impact of headers on response dynamics, within individual verbs we occasionally see a trend such that the ``wait'' header results in less targeting of the main clause (XLM-RoBERTa with ``has'' and ``would,'' DistilGPT with ``was'' and ``would,'' etc.), but for many verbs we see no difference, or even the opposite trend. On the whole, the impact of auxiliary verb identity is for most models much stronger than that of header.

\section{Discussion}

In this paper we have reported on a series of experiments testing sensitivity of pre-trained language models to dynamics involved in responding to an utterance in dialogue. We focus specifically on at-issueness and ellipsis, and find that models show clear sensitivity to the special status of embedded clauses, and general preference to target main clause content---but they show mixed results in terms of understanding the interaction of response headers with targeting of at-issue versus not-at-issue content. Furthermore, they show certain basic limitations in their grasp of the principles governing ellipsis, with selection of auxiliaries often influenced by superficial frequency factors rather than principled discourse constraints. Our findings also highlight differences between models, with certain models showing strong preference to target main clause content, and others showing stronger fluctuations based on individual auxiliary verbs.

This work highlights potential for improvement in standard PLMs, with respect to discourse sensitivities that have real implications for language competence generally, and for dialogue in particular. The models' sensitivity to special status of embedded clauses is consistent with work indicating sensitivity to syntax in these models~\citep{goldberg2019assessing}, and the consistency with which BERT and RoBERTa prefer auxiliaries targeting main clause content indicates that these models pick up on some interaction between ellipsis and syntax of previous context. Additionally, the slight impacts of header in Section~\ref{sec:rejection} suggest that these models may pick up on the beginnings of a relationship between response types and the types of content that they target. However, the general weakness in sensitivity to headers, failure on many aspects of the ellipsis tests, and interference of superficial factors, indicate clear room for growth in capturing the full range of these discourse dynamics. 

From a perspective of downstream dialogue tasks, our findings indicate that discourse competence in standard PLMs is not sufficiently comprehensive to expect that these models can provide a fully robust foundation for dialogue applications. It is possible---though not guaranteed---that training or fine-tuning directly for dialogue could improve the robustness of models' sensitivity to the specific types of response dynamics tested for here. We leave this question for future work.

\section*{Acknowledgments}

We are grateful to Ming Xiang, Shane Steinert-Threlkeld, Tal Linzen, Najoung Kim, and members of the UChicago CompLing Lab, for valuable comments and discussion. We thank Kanishka Misra in particular for guidance on usage of the minicons library. We also thank three anonymous reviewers for their thoughtful feedback and suggestions. 

\bibliography{main.bib}
\bibliographystyle{acl_natbib}

\appendix

\section{Appendix: Error analysis on ellipsis and auxiliary preference}

\begin{figure}[h]
\center
\vspace{-0.5cm}
\includegraphics[width=0.5\textwidth, height=5.4cm]{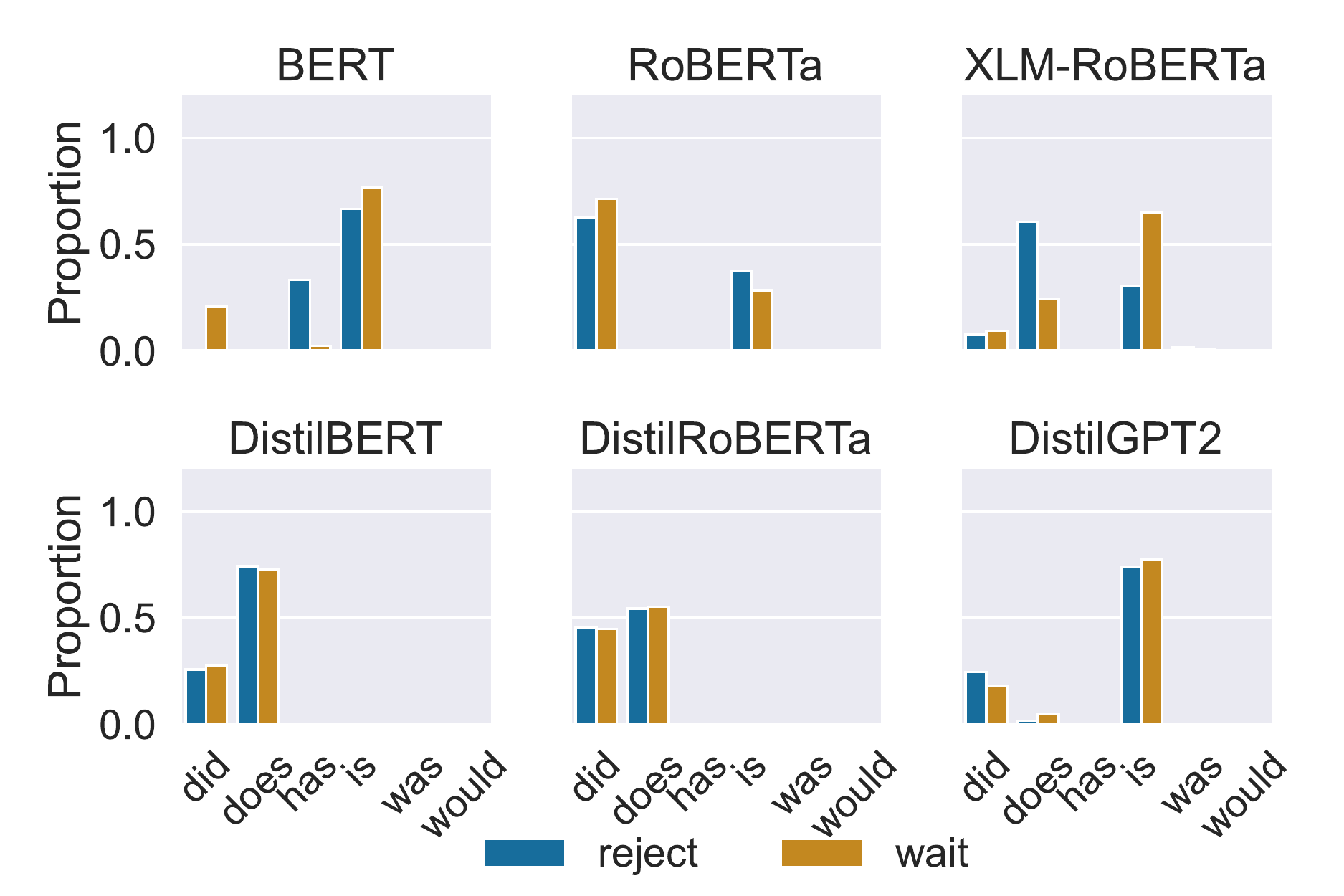}
\caption{\label{ellipsis-errorProportion-top1}Verb analysis on ellipsis test (top-1). The Y-axis indicates the proportion of the verb appearing as the top-1 prediction even when the corresponding auxiliary verb did not appear in the input sequence. Proportion is calculated by header. The X-axis shows the auxiliary verb used in the target sentence. `Reject' = ``No'' header; `Wait' = ``Wait no'' header.}
\end{figure}

\begin{figure}[h]
\center
\includegraphics[width=0.5\textwidth, height=5.4cm]{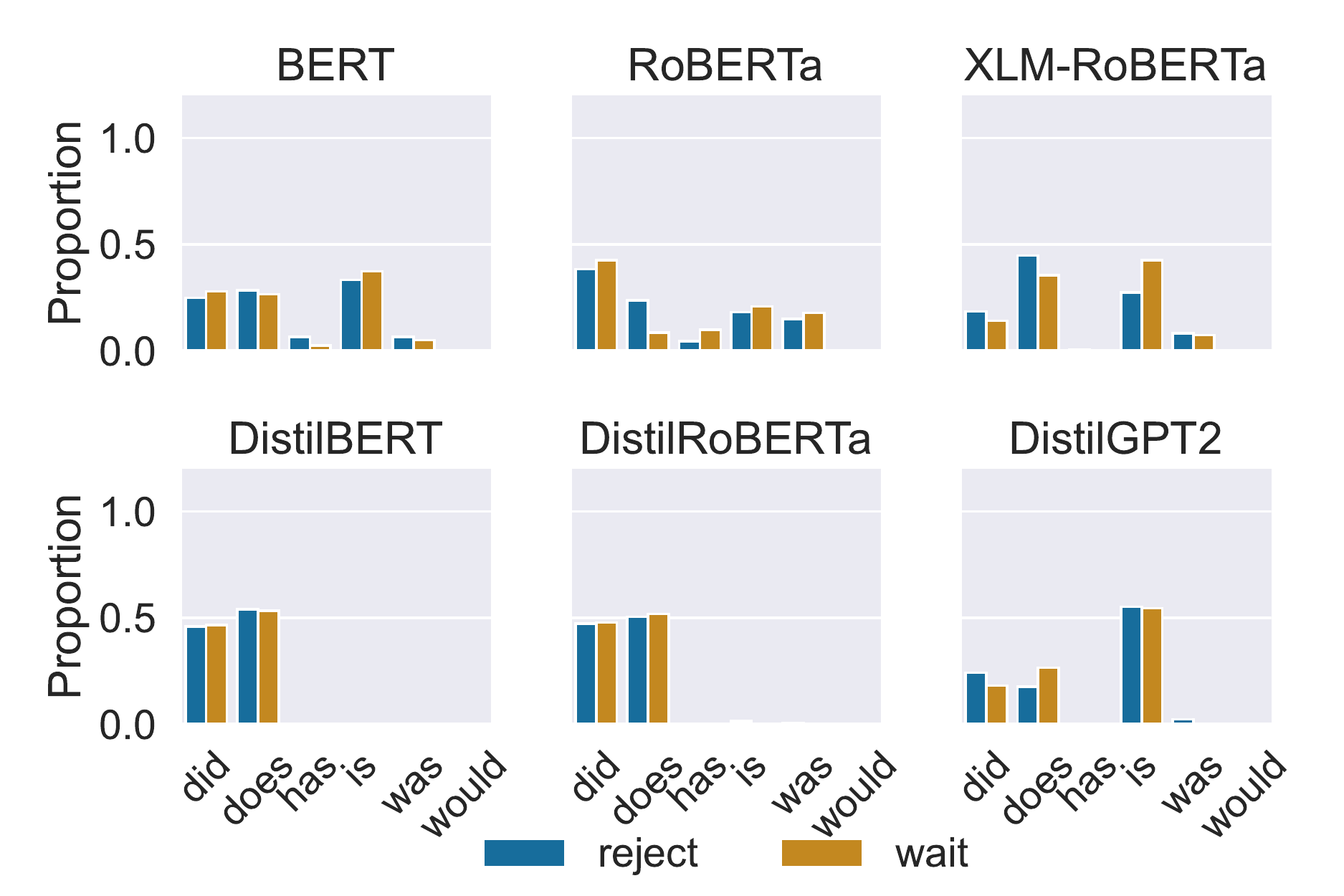}
\caption{\label{ellipsis-errorProportion-top2}Verb analysis on ellipsis test (top-2). The Y-axis indicates the proportion of the verb appearing as one of top-2 predictions even when the corresponding auxiliary verb did not appear in the input sequence. Proportion is calculated by header. The X-axis shows the auxiliary verb used in the target sentence. `Reject' = ``No'' header; `Wait' = ``Wait no'' header.}
\vspace{7cm}
\end{figure}

\end{document}